\documentclass[11pt,a4paper]{article}
\usepackage{times,latexsym}
\usepackage{url}
\usepackage[T1]{fontenc}
\usepackage[dvipsnames]{xcolor}
\usepackage{times}
\usepackage{latexsym}
\usepackage{iitem}
\usepackage[utf8]{inputenc}
\usepackage{microtype}
\usepackage{graphicx}
\usepackage{graphics}
\usepackage{subfigure}
\usepackage{booktabs} 
\usepackage{comment}
\usepackage{colortbl}
\usepackage{arydshln}
\usepackage{makecell}
\usepackage{hyperref}
\usepackage{mathtools}
\usepackage{color}
\usepackage{CJKutf8}
\usepackage{url}
\usepackage{multicol,multirow}
\usepackage{makecell}
\usepackage{array}
\usepackage{bm}
\usepackage{pifont}
\usepackage{parskip}
\usepackage{booktabs}       
\usepackage{amsmath, amsthm}
\usepackage{amssymb}
\usepackage{amsfonts}
\usepackage{array}

\usepackage{enumitem}
\usepackage{microtype}

\usepackage[acceptedWithA]{tacl2021v1}

\usepackage{xspace,mfirstuc,tabulary}

\newif\iftaclinstructions
\taclinstructionsfalse 
\iftaclinstructions
\renewcommand{\confidential}{}
\renewcommand{\anonsubtext}{(No author info supplied here, for consistency with
TACL-submission anonymization requirements)}
\newcommand{\instr}
\fi

\iftaclpubformat 

\else

\fi

\newcommand{\sts}{{{\textsc{Seq2Seq}}}\xspace}
\newcommand{\heart}{\text{\small \ding{170}}}

\title{Sentence Similarity Based on Contexts}
\date{}

\author{Xiaofei Sun$^\clubsuit$, Yuxian Meng$^\clubsuit$,  Xiang Ao$^\blacktriangle$, Fei Wu$^\blacklozenge$\\
{\bf Tianwei Zhang}$^\heart$, 
{\bf Jiwei Li}$^{\blacklozenge\clubsuit}$ and {\bf Chun Fan}$^{\spadesuit\bigstar}$\\
  $^\blacklozenge$Zhejiang University,
  $^\spadesuit$ Peking University,
  $^\bigstar$Peng Cheng Laboratory\\
  $^\blacktriangle$ Chinese Academy of Sciences,  $^\heart$ Nanyang Technological University
 \\
  $^\clubsuit$Shannon.AI\\
  \{xiaofei\_sun, yuxian\_meng, jiwei\_li\}@shannonai.com, aoxiang@ict.ac.cn\\
  tianwei.zhang@ntu.edu.sg, 
  fanchun@pku.edu.cn,
  wufei@zju.edu.cn
}

\begin{document}
\maketitle

\begin{abstract}
Existing methods to measure sentence similarity are faced with two challenges: (1) labeled datasets are usually limited in size, making them  insufficient to train supervised neural models; (2) there is a  training-test gap for unsupervised language modeling (LM) based models to compute semantic scores between sentences, since sentence-level semantics are not explicitly modeled at training. This results in inferior performances in this task. 
In this work, we propose a new framework  to address  these two issues. The proposed framework is based on the core idea that the meaning of a sentence should be defined by its contexts, and that sentence similarity can be measured by comparing the probabilities of generating two sentences given  the same context. The proposed framework is able to  generate  high-quality, large-scale dataset with semantic similarity scores between two sentences in an unsupervised manner, with which the train-test gap can be largely bridged. Extensive experiments show that the proposed framework achieves significant performance boosts over existing baselines under both the supervised and unsupervised settings across different datasets.\footnote{Accepted by TACL.}
\end{abstract}

\section{Introduction}
Measuring sentence similarity is a long-standing task in NLP \citep{luhn1957statistical,robertson1995okapi,blei2003latent,10.1145/3366423.3379998}. The task aims at quantitatively measuring the semantic relatedness between two sentences, and has wide applications in text search \citep{farouk2018graph}, natural language understanding \citep{maccartney2009natural} and machine translation \citep{yang2019sentence}.

One of the greatest challenges that existing methods face for sentence similarity is the 
lack of  large-scale labeled datasets, which contain sentence pairs with labeled semantic similarity scores.
The acquisition of such dataset is 
 both labor-intensive and expensive. For example, the STS benchmark \citep{cer2017semeval} and SICK-Relatedness dataset \citep{marelli-etal-2014-sick} 
 respectively contain 8.6K and 9.8K labeled sentence pairs, the sizes of which are usually insufficient for training deep neural networks.

Unsupervised learning methods are proposed to address this issue, where word embeddings \citep{le2014distributed} or BERT embeddings \citep{devlin2018bert} are used to 
to map sentences to fix-length vectors in an unsupervised manner. 
Then sentence similarity is computed based on the cosine or dot product of these sentence representations. 
Our work follows this thread where sentence similarity is computed  based on fix-length sentence  representations, 
as opposed to comparing sentences directly.  
The biggest issue with current unsupervised approaches is that 
there exists a big gap between model training and testing (i.e., computing semantic similarity between two sentences). 
For example, the BERT-style models are trained at the token level by predicting words given contexts, and there is neither explicit modeling sentence semantics nor producing sentence embeddings at the training stage.  
But at test time, sentence semantics needs to be explicitly modeled to obtain semantic similarity. 
The inconsistency results in a distinct discrepancy between the objectives at the two stages and inferior performances on textual semantic similarity tasks.
For example, BERT embeddings yield inferior performances on semantic similarity benchmarks \cite{reimers2019sentence}, and even underperforming the naive method such as averaging GloVe \citep{pennington2014glove} embeddings. 
\citet{li-etal-2020-sentence} investigated this problem and  found that BERT always induces a non-smooth anisotropic semantic space of sentences, and this property significantly harms the performance of semantic similarity. 

Like word meanings are defined by neighboring words \citep{harris1954distributional}, the meaning of a sentence is determined by its contexts. 
Given the same context, it is a high probability to generate two similar sentences. If it is a low probability of generating two sentences given the same context, there is a gap between these two sentences in the semantic space. 
Based on this idea, we propose a framework that measures semantic similarity  through  the probability similarity of generating two sentences given the same context in a fully unsupervised manner. 
As for implementation,
the framework consists of the following steps: 
(1)  we train a contextual model  by predicting the probability of a sentence fitting into the left and right contexts;
(2) we obtain sentence pair similarity  by comparing  scores assigned by the contextual model across a large number of contexts.
To facilitate inference, we train a surrogate model, to act as the role of step 2, based on the outputs from step 1.
The surrogate model can be directly used for sentence similarity prediction in an unsupervised setup, or used as initialization to be further finetuned on downstream datasets in the supervised setup. 
Note that the outcome from step 1 or the surrogate model is a fixed-length vector regarding the input sentence. Each element in the vector indicates how fit the input sentence is to the context corresponding to that element, and the vector itself can be viewed as the overall semantics of the input sentence in the contextual space. 
Then we use cosine distance between two sentence vectors to compute the semantic similarity.

The proposed framework offers the potential to fully address the two challenges above: 
(1) the context regularization provides a reliable means to generate a large-scale high-quality dataset with semantic similarity scores based on unlabeled corpus; and (2) the train-test gap can be naturally bridged by training the model on the large-scale similarity dataset, leading to significant performance gains compared to utilize pretrained models directly. 

We conduct experiments on different datasets under both supervised and unsupervised setups, and experimental results show that the proposed framework significantly outperforms existing sentence similarity models. 

\section{Related Work}
Statistics-based 
 methods for measuring sentence  similarity include 
    bag-of-words (BoW) \citep{li2006sentence}, term frequency inverse document frequency (TF-IDF) \citep{luhn1957statistical,jones1972statistical}, BM25 \citep{robertson1995okapi}, latent semantic indexing (LSI) \citep{deerwester1990indexing} and latent dirichlet allocation (LDA) \citep{blei2003latent}. Deep learning based methods for sentence similarity rely on distributed representations \citep{mikolov2013distributed,le2014distributed} and can be generally divided into the following three categories.

\paragraph{Matrix Based Methods}~{}\newline
The first line of work for measuring sentence similarity is to construct a similarity matrix between two sentences, each element of which represents the similarity between the two corresponding units in two sentences. Then the matrix is aggregated in different ways to induce the final similarity score.
\citet{pang2016text} applied a two-layer convolutional neural network (CNN) followed by a feed-forward layer to the similarity matrix to derive the similarity score.
\citet{he-lin-2016-pairwise}  used a  deeper CNN  to make the best use of the similarity matrix.
\citet{yin-schutze-2015-multigrancnn} built a hierarchical architecture to model text compositions at different granularities, so several similarity matrices can be computed and combined for interactions.
Other works proposed to use the attention mechanism as a way of computing the similarity matrix \citep{rocktaschel2015reasoning,wang-etal-2016-sentence,parikh2016decomposable,seo2016bidirectional,shen2017inter,lin2017structured,gong2017natural,tan2018multiway,kim2019semantic,yang-etal-2019-simple}.

\paragraph{Word Distance Based Methods}~{}\newline
The second line of work to measure sentence similarity is to calculate the cost of transforming from one sentence to another, and the smaller the cost is, the more similar two sentences are. This idea is  implemented by the Word Mover's Distance (WMD) \citep{kusner2015word},
which measures the dissimilarity between two documents as the minimum amount of distance that the embedded words of one document need to transform to words of another document. 
Following works improve WMD by incorporating supervision from downstream tasks \citep{huang2016supervised},  introducing hierarchical optimal transport over topics \citep{yurochkin2019hierarchical}, 
addressing the complexity limitation of requiring to consider each pair \citep{wu2017topic,wu2018word,backurs2020scalable} and 
combining graph structures with WMD to perform cross-domain alignment \citep{chen2020graph}.
More recently, \citet{yokoi-etal-2020-word} proposes to disentangle word vectors in WRD has shown significantly performance boosts over vanilla WMD.

\paragraph{Sentence Embeddings Based Methods}~{}\newline
Sentence embeddings are high-dimensional representations for sentences. They are expected to contain rich sentence semantics so that the similarity between two sentences can be computed by considering their sentence embeddings via certain metrics such as cosine similarity.
\citet{le2014distributed} introduced paragraph vector, which is learned in an unsupervised manner by predicting the words within the paragraph using the paragraph vector.
In a followup, a line of sentence embedding methods such as FastText, Skip-Thought vectors \citep{kiros2015skip}, Smooth Inverse Frequency (SIF) \citep{arora2016simple}, Sequential Denoising Autoencoder (SDAEs) \citep{hill-etal-2016-learning-distributed}, InferSent \citep{conneau2017supervised}, Quick-Thought vectors \citep{logeswaran2018efficient} and Universal Sentence Encoder \citep{cer2018universal}
have been proposed to improve the sentence embedding quality with more efficiency. 

The great success achieved by large-scale pretraining models \citep{devlin2018bert,yinhan2019roberta} has recently stimulated a strand of work on producing sentence embeddings based on the pretraining-finetuning paradigm using large-scale unlabeled corpora. The cosine outcome between the representations of two sentences produced by large-scale pretrained models is treated as the semantic similarity \citep{reimers2019sentence,wang2020sbert,li-etal-2020-sentence}. 
\citet{su2021whitening,huang2021whiteningbert} proposed to regularize the sentence representations by whitening them, i.e., enforcing the covariance to be an identity matrix to address the non-smooth anisotropic distribution issue \citep{li-etal-2020-sentence}.

The BERT-based scores \citep{Zhang2020BERTScore,sellam-etal-2020-bleurt}, though serve as automatic metrics, also capture rich semantic information regarding the sentence and have the potentials for measuring semantic similarity. \citet{cer2018universal} proposed a method of encoding sentences into their corresponding embeddings that specifically target transfer learning to other NLP tasks. \citet{karpukhin-etal-2020-dense} adopted two unique BERT encoder models and  the model weights are optimized to maximize the dot product. 
The most recent line of work focuses on leveraging the contrastive learning framework to tackle semantic textual similarity \citep{wu2020clear,carlsson2021semantic,kim-etal-2021-self,yan-etal-2021-consert,gao2021simcse}, where two similar sentences are pulled close and two random sentences are pulled away in the sentence representation space. This learning strategy helps better separate sentences with different semantics.

This work is motivated by learning word representations given its contexts \citep{mikolov2013distributed,le2014distributed} with the assumption that the meaning of a word is determined by its context. Our work is based on large-scale pretrained model and aims at learning informative sentence representations for measuring sentence similarity. 

\section{Model}
\subsection{Overview}

The key point of the proposed paradigm is to compute semantic similarity between two sentences by measuring the probabilities of generating the two sentences across a number of context. 

We can 
achieve this goal based on the following steps: 
(1) 
 we first need to 
train a contextual model to predict the probability of a sentence fitting into the left and right contexts. This goal can be achieved by either a discriminative model, i.e., predicting the probability that the concatenation of a sentence with context forms a coherent text, or a generative model, i.e., predicting the probability of generating a sentence given contexts;
(2) next, given a pair of sentences, we can measure their similarity by comparing their scores assigned by contextual models given different contexts; (3) for step 2, for any pair of sentences at test time, we need to sample different contexts to compute scores assigned by 
contextual models, which is time-consuming. We thus propose to train a surrogate model that takes a pair of sentences as inputs, and predicts the similarity assigned by the contextual model. 
This enables faster inference though at a small sacrifice of accuracy; 
(4) the surrogate model can be directly used for obtaining sentence similarity scores in a unsupervised manner, or used as model initialization, which will be further fine-tuned on downstream datasets in a supervised setting. 
We will discuss the detail of each module in order below. 

\subsection{Training Contextual Models}
 We  need  a contextual model to predict the probability of a sentence fitting into left and right contexts. 
We combine a generative model and a discriminative model to achieve this goal, allowing us to take the advantage of both to model text coherence \citep{li2017adversarial}.  
\paragraph{Notations}
Let $\bm{c}_i$ denote the $i$-th sentence, which consists of a sequence of words  $\bm{c}_i = \{{c}_{i,1}, ..., {c}_{i,n_i}  \}$, where $n_i$ denotes the number of words in $\bm{c}_i$.
Let 
$\bm{c}_{i:j}$ denote the $i$-th to $j$-th sentences. $\bm{c}_{<i}$ and $\bm{c}_{>i}$ respectively denote the preceding and subsequent context of $\bm{c}_i$.
\subsubsection{Discriminative Models}
\label{model:dis}
The discriminative model takes a sequence of consecutive sentences $[\bm{c}_{<i}, \bm{c}_{i}, \bm{c}_{>i}]$ as the input, and maps the input to a probability indicating whether the input is natural and coherent. 
We treat sentence sequences taken from the original articles written by humans as  positive examples and sequences with replacements of the center sentence $\bm{c}_{i}$ as negative ones.
Half of replacements of 
 $\bm{c}_{i}$ come from the original document, and half of 
replacements come from random sentences from the corpus. 
 The concatenation of LSTM representations at the last step (right-to-left and left-to-right) is used to represent the sentence. 
      Sentence representations  for consecutive sentences are concatenated and output to the sigmoid function to obtain the final probability:
\begin{equation}
p(y=1 | {\bm{c}}_i, \bm{c}_{<i},\bm{c}_{>i}) =\text{sigmoid}(\bm{h}^\top [\bm{h}_{<i}, \bm{h}_i, \bm{h}_{>i}]) 
\label{dis}
\end{equation}
where $\bm{h}$ denotes learnable parameters.
We deliberately make the discriminative model simple for two reasons: the discriminative approach for coherence prediction is a relatively easy task and more importantly, 
it will be further used in the next selection stage for screening, where faster speed is preferred. 

\subsubsection{Generative Models}
\label{model:gen}
Given  contexts $\bm{c}_{<i}$ and  $\bm{c}_{>i}$, 
the generative model predicts the probability of generating 
each token in 
sentence $\bm{c}_i$ sequentially using \sts structures \citep{sutskever2014sequence} as the backbone:
\begin{equation}
p(\bm{c}_i|\bm{c}_{<i},\bm{c}_{>i}) = \prod_{j} p(\bm{c}_{i,j}|\bm{c}_{<i},\bm{c}_{>i}, \bm{c}_{i,<j}) 
\end{equation}
Semantic similarity between two sentences can  be measured by
not only
 the forward probability of 
 generating the two sentences given the same context $p(\bm{c}_i|\bm{c}_{<i},\bm{c}_{>i})$, 
 but  also  the backward probability of generating contexts given sentences. 
The context-given-sentence probability can be modeled by 
  predicting preceding contexts given subsequent contexts $p(\bm{c}_{<i}|\bm{c}_i,\bm{c}_{>i})$ and to predict subsequent contexts given preceding contexts $p(\bm{c}_{>i}|\bm{c}_{<i},\bm{c}_i)$. 


\subsection{Scoring Sentence Pairs}
\label{scoring}
Given context $[\bm{c}_{<i}, \bm{c}_{>i}]$, the score for $\bm{s}_i$ fitting into the context is the linear combination of 
scores from discriminative and generative models:
\begin{equation}
\small
\begin{aligned}
&S(\bm{s}_i, \bm{c}_{<i}, \bm{c}_{>i}) =  \lambda_1 \log p(y=1 | \bm{s}_i, \bm{c}_{<i},\bm{c}_{>i}) \\
&+ \lambda_2 \frac{1}{|\bm{s}_i|}\log p(\bm{s}_i|\bm{c}_{<i},\bm{c}_{>i}) \\
& + \lambda_3 \frac{1}{|\bm{c}_{<i}|}\log p(\bm{c}_{<i}| \bm{s}_i,\bm{c}_{>i}) + \lambda_4 \frac{1}{|\bm{c}_{>i}|}\log p(\bm{c}_{>i}|\bm{c}_{<i},\bm{s}_i)
\end{aligned}
\label{score}
\end{equation}
where $\lambda_1$,  $\lambda_2$, $\lambda_3$,  $\lambda_4$ control the tradeoff between different modules.
For simplification, we  use $\bm{c}$ to denote context $\bm{c}_{<i}, \bm{c}_{>i}$.
$S(\bm{s}_i, \bm{c})$ is thus equivalent to 
$S(\bm{s}_i, \bm{c}_{<i}, \bm{c}_{>i})$.

Let $\bm{C}$ denote a set of  contexts, where $N_{\bm{C}}$ is the size of $\bm{C}$. 
For a sentence $\bm{s}$, its semantic representation $\bm{v}_{\bm{s}}$ is an $N_{\bm{C}}$ dimensional vector, with each individual  value being $S(\bm{s}, \bm{c}) $ with $\bm{c}\in \bm{C}$. 
The  semantic similarity between two sentences $\bm{s}_1$ and $\bm{s}_2$ can be computed based on $\bm{v}_{\bm{s}_1}$ and $\bm{v}_{\bm{s}_2}$ using different metrics such as cosine similarity. 

\paragraph{Constructing $\bm{C}$}
We need to pay special attentions to the construction of $\bm{C}$. 
The optimal situation is to use all contexts, where $\bm{C}$  is the entire corpus. 
Unfortunately, this is computationally prohibitive as we need to iterate over the entire corpus for each sentence $\bm{s}$. 

We propose the following workaround for tractable computation.  
For a sentence $\bm{s}$,
rather than using the full corpus as $\bm{C}$, 
 we construct its sentence specific context set $\bm{C}_{\bm{s}}$
in a way that $s$ can fit into all  constituent context in $\bm{C}_{\bm{s}}$. 
The intuition is as follows:   with respect to  sentence $\bm{s}_1$,  
contexts can be divided into two categories:
contexts which $\bm{s}_1$ fits into, based on which we will measure whether or not $\bm{s}_2$ also fits in;
contexts which $\bm{s}_1$ does not fit into, and we will measure whether or not $\bm{s}_2$ also does not fit in. 
We are mostly concerned about the former, and can neglect the latter.
The reason is as follows:
the latter can also further be divided into two categories:
contexts that  fit neither $\bm{s}_1$ or $\bm{s}_2$, and contexts that do not fit $\bm{s}_1$ but fit $\bm{s}_2$. 
For contexts that  fit neither $\bm{s}_1$ and $\bm{s}_2$, we can neglect them 
since two sentences not fitting into the same context does not signify their semantic relatedness;
for contexts that does not fit $\bm{s}_1$ but fit $\bm{s}_2$,  
we can leave them to when we compute $C_{\bm{s}_2}$.

Practically, for a given sentence $\bm{s}$, we first use a TF-IDF weighted bag-of-word bi-gram vectors to perform primary screening
on the whole corpus to retrieve related text chunks (20K for each sentence). 
Next, we rank all contexts 
 using the discriminative model based on Eq.\ref{dis}. 
 For discriminative models, we  cache sentence representations in advance, and compute 
model scores in the last neural layer, which is significantly faster than the generative model.
This two-step selection strategy is akin to the pipelined selection system \citep{chen2017reading,karpukhin-etal-2020-dense}
 in open-domain QA which contains document retrieval using IR systems and fine-grained question answering using neural QA models. 

$\bm{C}_{\bm{s}}$ is built by selecting top ranked contexts by Eq. \ref{score}. We use the incremental construction strategy, adding one context at a time.
To promote diversity of $\bm{C}_{\bm{s}}$, each text chunk is allowed to contribute at most one context, and the Jaccard similarity between 
the $i-1$-th sentence in the context to select and those already selected should be lower than 0.5.\footnote{This strategy can also remove text duplicates.} 

To compute semantic similarity between $\bm{s}_1$ and $\bm{s}_2$, we concatenate $\bm{C}_{\bm{s}_1}$ and $\bm{C}_{\bm{s}_2}$ and use the concatenation as the context set $\bm{C}$. 
The semantic  similarity score between $\bm{s}_1$ and $\bm{s}_2$ is given as follows:
\begin{equation}
\begin{aligned}
&\bm{v}_{\bm{s}_1} = [S(\bm{s}_1, \bm{c})\;\;\text{for }\bm{c}\in \bm{C}_{\bm{s}_1} + \bm{C}_{\bm{s}_2}] \\
&\bm{v}_{\bm{s}_2} = [S(\bm{s}_2,\bm{c})\;\;\text{for }\bm{c}\in  \bm{C}_{\bm{s}_1} + \bm{C}_{\bm{s}_2}] \\
&\text{sim}(\bm{s}_1, \bm{s}_2) = \text{cosine} (\bm{v}_{\bm{s}_1} , \bm{v}_{\bm{s}_2} )
\end{aligned}
\label{sim}
\end{equation}

\subsection{Training Surrogate Models}
\label{model:sur}
The method described in Section \ref{scoring} provides a direct way to compute scores for semantic relatedness.
But it comes with a severe shortcoming of slow speed at inference time:
given an arbitrary pair of sentences, the model still needs to go through the entire corpus, harvest the context set $\bm{C}_{\bm{s}}$, and iterate all instances in $\bm{C}_{\bm{s}}$ 
for context score calculation based on
 Eq.(\ref{score}), each of which is time consuming. 
To address this issue, we propose to train a surrogate model to accelerate inference.

Specifically, we first harvest similarity scores for sentence pairs using methods in Section \ref{scoring}.
We collect scores for 100M pairs in total, which are further split into train/dev/test by 98/1/1. 
Next, by treating harvested similarity scores as gold labels, 
we train a neural model that takes a pair of sentence as an input, and predicts its similarity score.
The cosine similarity between the two sentence representations is the predicted semantic similarity, 
and we minimize the $L_2$ distance between predicted and golden similarities. 
The Siamese structure makes it possible that fixed-sized vectors for input sentences can be derived and stored, allowing for fast semantic similarity search, which we will discuss in detail in the ablation study section.


It is worth noting  both the advantages and  disadvantages  of  the surrogate model. 
For advantages, firstly, it can significantly speed up inference as it avoids the time-consuming process of iterating over the entire corpus to construct $C$.
Secondly, the surrogate shares the same structure with existing widely-used models such as BERT and RoBERTa, and can thus later  be easily finetuned on the human-labeled datasets in supervised learning; on the other hand, the origin model in Section \ref{scoring}  cannot be readily combined with other human-labeled datasets.
For disadvantages, the surrogate model inevitably comes with a cost of accuracy, as its upper bound is  the origin model in Section \ref{scoring}. 

\section{Experiments}
\subsection{Experiment Settings}
We evaluate the {\it Surrogate} model on Semantic Textual Similarity (STS), Argument Facet Similarity (AFS) corpus~\citep{misra-etal-2016-measuring}, and Wikipedia Sections Distinction \citep{ein-dor-etal-2018-learning} tasks. 
We perform both unsupervised and supervised evaluations on these tasks.
For unsupervised evaluations, models are directly used for obtaining sentence representations. 
For supervised evaluations, we use the training set to fine-tune all models and use the $L_2$ regression as the objective function. 
Additionally, we also conduct partially supervised evaluation on STS benchmarks. 

\paragraph{Implementation Details}
For discriminative model in \ref{model:dis}, we use a single-layer bi-directional LSTM as the backbone with the size of hidden states set to 300. 

For generative model in \ref{model:gen}, We implement the above three models, i.e. $p({\bm{c}}_i|\bm{c}_{<i},\bm{c}_{>i})$, $p(\bm{c}_{<i}| \bm{c}_i,\bm{c}_{>i})$ and 
  $p(\bm{c}_{>i}|\bm{c}_{<i},\bm{c}_i)$ based on the \sts structure, and use Transformer-large as the backbone \cite{vaswani2017attention}.
Sentence position embeddings and token position embeddings are added to word embeddings. The model is trained on a corpus extracted from CommonCrawl which contains 100B tokens. 

For the surrogate model in \ref{model:sur}, we use RoBERTa~\citep{yinhan2019roberta} as the backbone, and adopts the Siamese structure \citep{reimers2019sentence}, where two sentences are first mapped to vector representations using RoBERTa. We use the average pooling on the last RoBERTa layer to obtain the sentence representation. During training, we use Adam \citep{kingma2014adam} with learning rate of 1e-4, $\beta_1$ = 0.9, $\beta_2$ = 0.999. The trained surrogate model obtains an average $L_2$  distance of $7.4\times 10^{-4}$ on dev set when trained from scratch, and $6.1\times 10^{-4}$ when initialized using the RoBERTa-large model \citep{yinhan2019roberta}. 
We set $\bm{C}_{\bm{s}}$ to 500. 
 
\paragraph{Baselines}
We use the following models as baselines:
\begin{itemize}[noitemsep]
  \item {\bf Avg. Glove embeddings} is the average of word embeddings produced via the co-occurrence statistics in the corpus \citep{pennington2014glove}.
    \item {\bf Avg. Skip-Thought embeddings} is the average of word embeddings produced by Skip-Thought vectors \citep{kiros2015skip}.
   \item {\bf InferSent} uses a siamese BiLSTM network with max-pooling over the output on NLI datasets \citep{conneau2017supervised}. 
  \item {\bf Avg. BERT embeddings} is the average of word embeddings produced by BERT \citep{devlin2018bert}.
  \item {\bf BERT [CLS]} computes scores based on the vector representation of the special token [CLS] in BERT.
   \item {\bf BERTScore} computes the similarity of two sentences as a sum of cosine similarities between their tokens’ embeddings
   \citep{Zhang2020BERTScore}. 
  \item {\bf BLEURT} is baseed on BERT and captures non-trivial semantic similarities by fine-tuning the model on the WMT Metrics dataset, on a set of ratings provided by the user, or a combination of both~\cite{sellam-etal-2020-bleurt}. 
  \item {\bf DPR} works by using two unique BERT encoder models and  the model weights are optimized to maximize the dot product~\cite{karpukhin-etal-2020-dense}. 
  \item {\bf Universal Sent Encoder} is a method of encoding sentences into their corresponding embeddings that specifically target transfer learning to other NLP tasks \citep{cer2018universal}.
  \item {\bf SBERT} is a BERT-based method of using the Siamese structure to derive sentence embeddings that can be compared through cosine similarity \citep{reimers2019sentence}. 
\end{itemize}

\subsection{Run-time Efficiency}
\begin{table}[t]
    \centering
    \small
    \begin{tabular}{lcc}\toprule
    {\bf Model} &  {\bf  CPU} & {\bf GPU}    \\\midrule
    InferSent & 125  &  1527 \\
    Universal Sent Encoder  & 72  & 1330 \\
    SBERT$_{base}$ & 41 & 1315 \\
SBERT$_{base}$ length batching & 88 & 2112 \\ 
Surrogate$_{base}$ &  48   &   1514 \\
Surrogate$_{base}$ length batching & 91 &   2175 \\\bottomrule
    \end{tabular}
    \caption{ Computation speed of sentence embedding methods(sentences per second). }
    \label{exp:speed}
\end{table}
The run-time efficiency is important for sentence representation models  since similarity functions are potentially applied to large corpora. 
In this subsection, we compare {\it Surrogate$_{base}$} to InferSent~\citep{conneau2017supervised}, Universal Sent Encoder~\citep{cer2018universal} and SBERT$_{base}$~\citep{reimers2019sentence}. We adopt a length batching strategy in which 
sentences are grouped together by length.

The proposed {\it Surrogate} model is based on PyTorch. 
InferSent~\citep{conneau2017supervised} and SBERT~\citep{reimers2019sentence} are based on PyTorch.
Universal Sent Encoder~\citep{cer2018universal} is based on Tensorflow and the model is from the Tensorflow Hub.  
Model efficiency is measured on a server with Intel i7-5820K CPU @ 3.30GHz, Nvidia Tesla V100 GPU, CUDA 10.2 and cuDNN. 
We report both CPU and GPU speed and the results can be found in Table \ref{exp:speed}. As can be seen, InferSent is around 69\% faster than {\it Surrogate} model on CPU since its simpler model architecture. 
The speed
of the proposed  {\it Surrogate} model is comparable to {\it SBERT} for both non-batching and batching setups, which is in accord with our expectations due the same 
transformer structure adopted by the  {\it Surrogate} model. 

\begin{table*}[!t]
  \centering
  \small
  \scalebox{0.9}{
  \begin{tabular}{lcccccccc}\toprule
    {\bf Model} & {\bf STS12} & {\bf STS13} & {\bf STS14} & {\bf STS15} & {\bf STS16} & {\bf STSb} & {\bf SICK-R} & {\bf Avg}\\\midrule
    & \multicolumn{8}{c}{{\it fully unsupervised without human labels}}\vspace{2pt}\\
    {\it Avg. Glove embeddings$^\text{$\S$}$} & 55.14& 70.66& 59.73& 68.25 &63.66 &58.02& 53.76& 61.32 \\
    {\it Avg. Skip-Thought embeddings$^\text{$\S$}$} & 57.11 & 71.98& 61.30&70.13&65.21&59.42&55.50&62.95  \\
    {\it InferSent-Glove$^\text{$\sharp$}$} &
    52.86 & 66.75 & 62.15 & 72.77 & 66.87& 68.03&65.65 &65.01 \\
    {\it Avg. BERT embeddings$^\text{$\S$}$} & 38.78 &57.98& 57.98 &63.15& 61.06& 46.35 &58.40 &54.81\\
    {\it BERT [CLS]$^\text{$\sharp$}$} & 20.16 & 30.01&20.09& 36.88 & 38.08 & 16.50&42.63& 29.19  \\
    {\it BERTScore$^\text{$\sharp$}$} &54.60 & 50.11& 57.74 & 70.79 & 64.58 & 57.58 & 51.37&58.11 \\
    {\it DPR$^\text{$\sharp$}$}&53.98&56.00&57.83&66.68&67.43&58.53&61.85 &60.33\\
    {\it BLEURT$^\text{$\sharp$}$} & 70.16  & 64.97  & 57.41  & 72.91  & 70.01  & 69.81  & 58.46  & 66.25  \\
    {\it Universal Sent Encoder$^\text{$\sharp$}$} & 64.49 &67.80 &64.61 &76.83& 73.18 &74.92& 76.69 &71.22\\\cdashline{1-9}
    {\it Origin}  &  {\bf 72.41}    & {\bf 74.30}    &   {\bf 75.45}   &   {\bf 78.45}  &  {\bf 79.93} &{\bf 78.47} & {\bf 79.49} & {\bf 76.93}\\
    {\it Surrogate$_\text{base}$}  & 70.62  &  72.14   & 72.72   & 76.34   &   75.24   &74.19 &   77.20& 74.06\\
    {\it Surrogate$_\text{large}$}  & 71.93  &  73.74  &  73.95   &  77.01 &  76.64   & 75.32  &  77.84& 75.20\\\midrule
  & \multicolumn{8}{c}{{\it partially supervised without human labels but not the same domain}}\vspace{2pt}\\
   {\it InferSent-NLI$^\text{$\sharp$}$} & 50.48& 67.75 & 62.15 &72.77 &66.87& 68.03 & 65.65 &64.81 \\
  {\it BERT [CLS]-NLI$^\text{$\sharp$}$} & 60.35 & 54.97 & 64.92 & 71.49 & 70.49 & 73.25 & 70.79 & 66.61\\
{\it BERTScore-NLI$^\text{$\sharp$}$} & 60.89 & 54.64 & 63.96 & 74.35 & 66.67 & 65.65 & 66.01 & 64.60\\
{\it DPR-NLI$^\text{$\sharp$}$} & 61.36&56.71& 65.49& 71.80& 71.03&74.08 & 70.86 &67.33  \\
{\it BLEURT-NLI$^\text{$\sharp$}$}&66.40& 68.15&71.98 &79.69&77.86&77.98 & 70.92 & 73.28 \\ 
{\it Universal Sent Ecoder-NLI$^\text{$\sharp$}$} &65.55&67.95& 71.47&80.81&78.70&78.41&69.31&73.17
\\\cdashline{1-9} 
  {\it BERT-NLI$_\text{base}^\text{$\sharp$}$}&71.07&76.81&73.29&79.56& 74.58&77.10&72.65&75.01 \\
  {\it SBERT-NLI$_\text{base}^\text{$\S$}$} &70.97& 76.53 &73.19& 79.09 &74.30& 77.03& 72.91& 74.86\\
  {\it SRoBERTa-NLI$_\text{base}^\text{$\S$}$} &71.54 &72.49 &70.80&78.74 &73.69& 77.77& {74.46}& 74.21\\
  {\it Surrogate-NLI$_\text{base}$} & 74.15 & 76.50 & 72.23 &   81.24    &      78.75   &  79.32 &   78.56& 77.25\\
  \cdashline{1-9}
  {\it BERT-NLI$_\text{large}^\text{$\sharp$}$}& 71.62& 77.40&72.69&78.61 &75.28 &77.83 &72.64 & 75.15 \\
  {\it SBERT-NLI$_\text{large}^\text{$\S$}$} &72.27 &{78.46} &{74.90}& 80.99 &76.25 &{79.23} &73.75& 76.55\\
  {\it SRoBERTa-NLI$_\text{large}^\text{$\S$}$} &{74.53}& 77.00& 73.18 &{81.85}& {76.82}& 79.10& 74.29& {76.68}\\
  {\it Surrogate-NLI$_\text{large}$}& {\bf 76.98}  &{\bf 79.83}  &{\bf 75.15}  &  {\bf 83.54}&{\bf 79.32}  & {\bf 80.82}  & {\bf 79.64}& {\bf 79.33}\\\bottomrule
  \end{tabular}
  }
  \caption{Spearman rank correlation $\rho$ between the cosine similarity of sentence representations and the gold labels for various Textual Similarity (STS) tasks under the unsupervised setting. We use *-NLI to denote the model additionally trained on NLI datasets. $\sharp$ indicates that results are reproduced by ourselves; $\S$ indicates results are taken from \citet{reimers2019sentence}; {\it Surrogate} are results for our proposed method. } 
  \label{tab:sts-unsupervised}
\end{table*}

\subsection{Experiment: Semantic Textual Similarity}
We evaluate the proposed method on the Semantic Textual Similarity (STS) tasks. We compute the Spearman's rank correlation $\rho$ between the cosine similarity of the sentence pairs and the gold labels for comparison.

\paragraph{Unsupervised Evaluation}
We evaluate the proposed method on the Semantic Textual Similarity (STS) tasks 2012 - 2016 \citep{10.5555/2387636.2387697,agirre-etal-2013-sem,agirre-etal-2014-semeval,agirre-etal-2015-semeval,agirre-etal-2016-semeval}, the STS benchmark \citep{cer2017semeval} and the SICK-Relatedness dataset \citep{marelli-etal-2014-sick} for evaluation. All datasets contain sentence pairs labeled between 0 and 5 as the semantic relatedness. 
The proposed models are directly used for inference under the unsupervised setup. 

The results are shown in Table \ref{tab:sts-unsupervised} and we observe significant performance boosts of the proposed models over baselines.
Notably, the proposed models trained in the unsupervised setting (both {\it Origin} and {\it Surrogate}) are able to achieve competitive results to models trained on additional annotated NLI datasets. 
Another observation is, as expected, the {\it Surrogate} models underperform the {\it Origin} model as {\it Origin} serves as an upper bound for {\it Surrogate} but with a cost of inference speed. 

\paragraph{Partially Supervised Evaluation}
We finetune the model on the combination of the SNLI \citep{bowman-etal-2015-large} and the Multi-Genre NLI \citep{williams-etal-2018-broad} dataset, with the former containing 570K sentence pairs and the latter containing 433K pairs across various genres of sources. Sentence pairs from both datasets are annotated with one of the labels  contradiction, entailment, and neutral. The proposed models are trained on the natural language inference task then used for computing sentence representations in an unsupervised manner. 

The partially supervised results are shown in Table \ref{tab:sts-unsupervised}. 
As can be seen, results from the proposed model finetuned on NLI datasets are comparable to results from unsupervised models since no labeled similarity dataset is used, 
and comparable to results from supervised models if further finetuned on similarity datasets such as STS. 

\paragraph{Supervised Evaluation}
For the supervised setting, we use the STS benchmark (STSb) to evaluate supervised STS systems.
This dataset contains 8,628 sentence pairs from three categories: captions, news, and forums, and  is split into 5,749/1,500/1,379 sentence pairs respectively for training/dev/test. The proposed models are finetuned on the labeled datasets under the setup. 

For our proposed framework, we use {\it Origin} to represent the original model, where $\bm{C}$ for each sentence is constructed by searching the entire corpus 
as in Section \ref{scoring}
and we compute similarity scores based on Eq.(\ref{sim}). 
We also report performances for {\it Surrogate} models with base and large sizes. 
 
The results are shown in Table \ref{tab:sts-supervised}. We can see that for both model sizes (base and large) and both setups (with and without NLI training), the proposed {\it Surrogate} model significantly outperforms baseline models, leading to an average of over 2-point performance gains on the STSb dataset.

Note that the  {\it Origin} model can not be readily adapted to the partially supervised or supervised setting because it is hard to finetune the {\it Origin} model where the context set $\bm{C}$ needs to be constructed first. Hence, we finetune the {\it Surrogate} model as a compensation for the accuracy loss brought by the replacement of {\it Origin} with {\it Surrogate}. As we can see from Table \ref{tab:sts-unsupervised} and Table \ref{tab:sts-supervised}, finetuning {\it Surrogate} on NLI datasets and STSb is an effective remedy for the performance loss.

\begin{table}[!t]
  \small
  \centering
  \scalebox{0.9}{
  \begin{tabular}{lc}
    \toprule 
    {\bf Model} & {\bf Spearman $\rho$}\\\midrule 
    {\it BERT [CLS]$^\text{$\sharp$}$} & 73.01 \\
    {\it BERT$_\text{base}^\text{$\S$}$}& 84.30\\
    {\it SBERT$_\text{base}^\text{$\S$}$}&84.67\\ 
    {\it SRoBERTa$_\text{base}^\text{$\S$}$} &84.92 \\
    {\it Surrogate$_\text{base}$} &  {\bf 87.91}\\
    \cdashline{1-2}
    {\it BERT-NLI$_\text{base}^\text{$\S$}$}& 88.33\\ 
    {\it SBERT-NLI$_\text{base}^\text{$\S$}$} &85.35\\
    {\it SRoBERTa-NLI$_\text{base}^\text{$\S$}$} &84.79\\
    {\it Surrogate-NLI$_\text{base}$} &{\bf 89.95}\\
    \midrule
    {\it BERT$_\text{large}^\text{$\S$}$}& 85.64\\
    {\it SBERT$_\text{large}^\text{$\S$}$}& 84.45\\
    {\it SRoBERTa$_\text{large}^\text{$\S$}$}& 85.02\\
    {\it Surrogate$_\text{large}$}&  {\bf 88.52}\\
    \cdashline{1-2}
    {\it BERT-NLI$_\text{large}^\text{$\S$}$} &88.77\\
    {\it SBERT-NLI$_\text{large}^\text{$\S$}$}& 86.10\\
    {\it SRoBERTa-NLI$_\text{large}^\text{$\S$}$}& 86.15\\
    {\it Surrogate-NLI$_\text{large}$} & {\bf 90.69}\\\bottomrule
  \end{tabular}
  }
  \caption{Spearman correlation $\rho$ for the STSb dataset under the supervised setting. We use *-NLI to denote the model additionally trained on NLI datasets. $\sharp$ indicates that results are reproduced by ourselves;  $\S$ indicates results are taken from \citet{reimers2019sentence}; {\it Surrogate} are results for our proposed method.}
  \label{tab:sts-supervised}
\end{table}

\begin{table}[!t]
  \small
  \centering 
  \scalebox{0.9}{
  \begin{tabular}{lcc}\toprule
    {\bf Model} & {\bf Pearson $r$} & {\bf Spearman $\rho$}\\\midrule 
    & \multicolumn{2}{c}{{\it Unsupervised Setting}}\vspace{2pt}\\
    {\it Avg. Glove embeddings$^\text{$\sharp$}$} & 32.40& 34.00\\
    {\it Avg. Skip-Thought embeddings$^\text{$\sharp$}$} &22.34&23.24  \\
    {\it InferSent-Glove$^\text{$\sharp$}$} & 24.83  & 25.83 \\
    {\it Avg. BERT embeddings$^\text{$\sharp$}$} &  29.15 & 31.45 \\
    {\it BERT [CLS]$^\text{$\sharp$}$} &12.00&9.06 \\
    {\it BERTScore$^\text{$\sharp$}$} & 45.32 & 33.56 \\
    {\it DPR$^\text{$\sharp$}$}  & 41.89 & 32.16  \\
    {\it BLEURT$^\text{$\sharp$}$} & 45.98 & 44.12  \\
    {\it Universal Sent Encoder$^\text{$\sharp$}$} &44.28&43.47 \\      \cdashline{1-3}
    {\it Origin} &  {\bf 56.20}  & 54.40\\
    {\it Surrogate$_\text{base}$}   &53.00  &  52.50\\
    {\it Surrogate$_\text{large}$}    & 54.50   & {\bf 54.70}\\\midrule 
    & \multicolumn{2}{c}{{\it Supervised Setting}}\vspace{2pt}\\
     {\it BERT [CLS]$^\text{$\sharp$}$} & 35.28&36.24\\
    {\it BERT$_\text{base}^\text{$\S$}$} &77.20 &74.84\\
    {\it SBERT$_\text{base}^\text{$\S$}$}& 76.57 &74.13\\
     {\it SRoBERTa$_\text{base}^\text{$\sharp$}$} & 77.26 & 74.89 \\
     {\it Surrogate$_\text{base}$} & 79.80 & 78.20\\
    \cdashline{1-3}
    {\it BERT$_\text{large}^\text{$\S$}$} &78.68& 76.38\\
    {\it SBERT$_\text{large}^\text{$\S$}$}&77.85 &75.93\\
     {\it SRoBERTa$_\text{large}^\text{$\sharp$}$} & 79.03 & 76.92 \\
    {\it Surrogate$_\text{large}$} &    {\bf 81.00}  &  {\bf 80.50}\\\bottomrule
  \end{tabular}
  }
  \caption{Results of Pearson correlation $r$ and Spearman's rank correlation $\rho$ on the Argument Facet
  Similarity (AFS) dataset. $\sharp$ indicates that results are reproduced by ourselves; $\S$ indicates results are taken from \citet{reimers2019sentence}; {\it Surrogate} are results for our proposed method.}
  \label{tab:afs}
\end{table}

\subsection{Experiment: Argument Facet Similarity}
We evaluate the proposed model on the Argument Facet Similarity (AFS) dataset \citep{misra-etal-2016-measuring}. This dataset contains 6,000 manually annotated argument pairs collected from human conversations on three topics: gun control, gay marriage and death penalty.  Each argument pair is labeled on a scale between 0 and 5 with a step of 1. Different from the sentence pairs in STS datasets, the similarity of an argument pair in AFS is measured not only in the claim, but also in the way of reasoning, which makes AFS a more difficult dataset compared to STS datasets. We report the Pearson correlation $r$ and Spearman's rank correlation $\rho$ to compare all models. 
\paragraph{Unsupervised Evaluation} 
The results are shown in Table \ref{tab:afs}, from which we can see for both the unsupervised settings, the proposed models {\it Origin} and {\it Surrogate} outperform baseline models by a large margin, with over 10 points for the unsupervised setting and over 4 points for the supervised setting. 

\paragraph{Supervised Evaluation}
We follow \citet{reimers2019sentence} to use the 10-fold cross-validation for supervised learning.
Results are shown in Table \ref{tab:afs}, from which we can see for both the supervised settings, the proposed models {\it Origin} and {\it Surrogate} outperform baseline models by a large margin, with over 10 points for the unsupervised setting and over 4 points for the supervised setting. 

\subsection{Experiment: Wikipedia Sections Distinction}
\citet{ein-dor-etal-2018-learning} constructed a large set of weakly labeled sentence triplets from Wikipedia for evaluating sentence embedding methods, each of which is composed of a pivot sentence, one sentence from the same section and one from another section. Test set contains 222K triplets.
The construction of this dataset is based on the idea that a sentence is thematically closer to sentences within its section than to sentences from other sections. 

We use accuracy as the evaluation metric for both unsupervised and supervised experiments: 
an example is treated as correctly classified if 
 the positive example is closer to the anchor than the negative example. 
\paragraph{Unsupervised Evaluation}
We directly evaluate the trained model on the test set without finetuning.  
Results are shown in Table \ref{tab:wiki}. For the unsupervised setting, the large model {\it Surrogate$_\text{large}$} outperforms the base model {\it Surrogate$_\text{base}$} by 2.1 points.

\paragraph{Supervised Evaluation}
During training, we use the triple objective to train the proposed model on 1.8M training triplets and evaluate it on the test set. 

Results are shown in Table \ref{tab:wiki}. For the supervised setting, the proposed model significantly outperforms SBERT, with a nearly 3-point gain in accuracy for both base and large models. 

\begin{table}[t]
  \small
  \centering 
  \begin{tabular}{lc}\toprule
    {\bf Model} &  {\bf Accuracy}\\\midrule 
    \multicolumn{2}{c}{{\it Unsupervised Setting}}\vspace{2pt}\\
    {\it Avg. Glove embeddings$^\text{$\sharp$}$} &60.94\\
    {\it Avg. Skip-Thought embeddings$^\text{$\sharp$}$} &61.54  \\
    {\it InferSent-Glove$^\text{$\sharp$}$} & 63.39\\
    {\it Avg. BERT embeddings$^\text{$\sharp$}$} &66.40   \\
    {\it BERT [CLS]$^\text{$\sharp$}$} & 32.30\\
    {\it BERTScore$^\text{$\sharp$}$} & 67.29 \\
    {\it DPR$^\text{$\sharp$}$}  &66.71\\
    {\it BLEURT$^\text{$\sharp$}$} & 67.39 \\
    {\it Universal Sent Encoder$^\text{$\sharp$}$} & 65.18\\
    {\it Surrogate$_\text{base}$}   &71.40  \\
    {\it Surrogate$_\text{large}$}    & 73.50  \\\midrule 
    \multicolumn{2}{c}{{\it Supervised Setting}}\vspace{2pt}\\
    {\it BERT [CLS]$^\text{$\sharp$}$} &78.13 \\
    {\it BERT$_\text{base}^\text{$\sharp$}$}&79.30 \\
    {\it SBERT$_\text{base}^\text{$\S$}$}& 80.42\\
    {\it SRoBERTa$_\text{base}^\text{$\S$}$}&79.45 \\
    {\it Surrogate$_\text{base}$} & 83.10 \\
    \cdashline{1-2}
     {\it BERT$_\text{large}^\text{$\sharp$}$}&80.15 \\
    {\it SBERT$_\text{large}^\text{$\S$}$}& 80.78\\
    {\it SRoBERTa$_\text{large}^\text{$\S$}$}& 79.73\\
    {\it Surrogate$_\text{large}$} & {\bf 83.50} \\\bottomrule
  \end{tabular}
  \caption{Accuracy results for the Wikipedia sections distinction task. $\sharp$ indicates that results are reproduced by ourselves; $\S$ indicates results are taken from \citet{reimers2019sentence}; {\it Surrogate} are results for our proposed method.}
  \label{tab:wiki}
\end{table}

\section{Ablation Studies}
We perform comprehensive ablation studies on the STSb dataset with no additional training on NLI datasets to better understand the behavior of the proposed framework. 
Studies are performed on both the original model setup (denoted by {\it Origin}) and the surrogate model setup (denoted by {\it Surrogate}). 
We adopt the unsupervised setting for comparison.

\subsection{Size of Training Data for {\it Origin}}
We would like to understand how the size of data for training {\it Origin} affects downstream performances. We vary the training size between [10M, 100M, 1B, 10B, 100B] and present the results in Table \ref{tab:size-origin}. The model performance drastically improves as we increase the size of training data when its size is below 1B. 
 With more training data, e.g. 1B and 10B, the performance is getting close to the best result achieved with the largest training data. 

\begin{table}[t]
  \small
  \centering 
  \begin{tabular}{cccccc}\toprule 
    {\it Size} & {\bf 10M} & {\bf 100B} & {\bf 1B} & {\bf 10B} & {\bf 100B} \\
    {\it Spearman $\rho$} &49.41 &  66.92  &76.17   & 77.81 &  {\bf 78.47}\\\bottomrule
  \end{tabular}
  \caption{The effect of size of training data for {\it Origin}.}
  \label{tab:size-origin}
\end{table}

\subsection{Size of $\bm{C_s}$}
Changing the size of $\bm{C_s}$ will  have an influence on downstream performances. Table \ref{tab:c} shows the results. The overall trend is clear: a larger $\bm{C}$ leads to better performances. When the size is 20 or 100, the results are substantially worse than the result when the size is 500. Increasing the size from 500 to 1000 only brings 
 marginal 
performance gains. We thus use 500 for a trade-off between performance and speed.

\subsection{Number of  Pairs to Train {\it Surrogate}}
Next, we would like to explore the effect of the number of sentence pairs to train {\it Surrogate}. The results are shown in Table \ref{tab:size-surrogate}. As expected, more training data leads to better performances. With only 100K training pairs, the {\it Surrogate}  model still achieves an acceptable result of 74.02, which indicates that the collected automatically labeled sentence pairs are of high quality.

\begin{table}[t]
  \small
  \centering 
  \begin{tabular}{ccccc}\toprule 
    {\it Size} & {\bf 20} & {\bf 100} & {\bf 500} & {\bf 1000} \\
    {\it Spearman $\rho$} &66.25  &  73.93   &  78.47   &{\bf 78.56}\\\bottomrule
  \end{tabular}
  \caption{The effect of size of $\bm{C}$.}
  \label{tab:c}
\end{table}

\begin{table}[t]
  \small
  \centering 
  \begin{tabular}{ccccc}\toprule 
    {\it Size} & {\bf 100K} & {\bf 1M} & {\bf 10M} & {\bf 100M} \\
    {\it Spearman $\rho$} &74.02 &  76.11 &   76.92  &{\bf 77.32}\\\bottomrule
  \end{tabular}
  \caption{The effect of training data size for {\it Surrogate}.}
  \label{tab:size-surrogate}
\end{table}

\subsection{How to Construct $\bm{C}$}
We explore the effect of the way we construct $\bm{C}$. 
We compare three different strategies: (1) the proposed two-step strategy as detailed in Section \ref{scoring}; (2) randomly selection; and (3) the proposed two-step strategy but without the diversity promotion constraint that allows each text chunk to contribute at most one context.
For all strategies, we fix the size of $\bm{C}$ to 500.

The results for these strategies are respectively 78.47, 34.45 and 76.32. The random selection strategy significantly underperforms the  other two. The explanation is as follows: given the huge semantic space for sentences, 
randomly selected contexts are very likely to be semantic irrelevant to both $\bm{s}_1$ and $\bm{s}_2$
and 
can hardly  reflect the contextual semantics the sentence resides in. The similarity computed using context scores based on completely irrelevant  contexts is thus extremely noisy, leading to 
inferior performances. 
Removing the diversity promotion constraint (the third strategy), 
 the Spearman correlation reduces by over 2 points.  The explanation is straightforward: without the diversity constraint, very similar contexts will be included in $\bm{C}$, making the dimensions in the semantic vector redundant; 
  with more diverse contexts, the sentence similarity can be measured more comprehensively and the result can be more accurate.

\subsection{Modules in the Scoring Function}
We next turn to explore the effect of each term  in the scoring function Eq.(\ref{score}). 
Table \ref{tab:module} shows the results. We can observe that removing each of these terms leads to performance drops to different degrees. Removing {\it discriminative} results in the least performance loss, with a reduction of 0.5; removing {\it left-context} and {\it right-context}  respectively  results in a performance loss of 1.11 and 1.46; and removing both  {\it left-context} and {\it right-context} has the largest negative impact on the final results, with a performance loss of 1.97. These observations verify the importance of different terms in the scoring function, especially the context prediction terms.

\begin{table}[t]
  \small
  \centering 
  \begin{tabular}{lc}\toprule 
    {\bf Model} & {\bf Spearman $\rho$}\\\midrule 
    {\it Full} & 78.47\\
    {\it w/o discriminative} & 77.97 (-0.50)\\
    {\it w/o left-context} & 77.36 (-1.11)\\
    {\it w/o right-context} & 77.01 (-1.46)\\
    {\it w/o both contexts} & 76.50 (-1.97)\\\bottomrule
  \end{tabular}
  \caption{The effect of each term in the scoring function Eq.(\ref{score}). {\it discriminative} stands for $\log p(y=1|\bm{s}_i, \bm{c}_{<i},\\\bm{c}_{>i})$, {\it left-context} stands for $\frac{1}{|\bm{c}_{<i}|}\log p(\bm{c}_{<i}| \bm{s}_i,\bm{c}_{>i})$ and {\it right-context}  stands for $ \frac{1}{|\bm{c}_{>i}|}\log p(\bm{c}_{>i}|\bm{c}_{<i},\bm{s}_i)$. {\it both contexts} means we remove both left context and right context.}
  \label{tab:module}
\end{table}
\subsection{Model Structures}
To train the surrogate model, we originally use the Siamese network structure where two sentences are separately feed into the same model. It would be interesting to see the effect of feeding two sentences together into the model, i.e., $\{\text{[CLS]}, \bm{s}_1, \text{[SEP]}, \bm{s}_2\}$ and then using the special token [CLS] for computing the similarity, which is the strategy that BERT uses for sentence-pair classification. Here,
we call it the BERT-style model for comparison with the Siamese model. 

By training the BERT-style model using the same harvested sentence pairs as the Siamese model with the $L_2$ regression loss, we obtain a Spearman's rank correlation of 77.43, 
slightly better than the 
 result of 77.32 for  the Siamese model.
This is because  interactions 
 between  words/phrases in two sentences are modeled more sufficiently in the BERT structure as interactions
start at the input layer through self-attentions. For the Siamese structure, the two sentences do not interact until the output cosine layer. 

The merit of sufficient interactions from the BERT structure also comes at a cost: we need to rerun 
the full model for any new sentence pair. This is not the case with the Siamese structure, which 
allows for  fast semantic similarity search by caching sentence representations in advance. 
In practice, we prefer the Siamese structure since the speedup in semantic similarity search overweighs the slight performance boost brought 
by the BERT structure. 
\subsection{Case Analysis}
We conduct case analysis on STS benchmark \citep{cer2017semeval} test set.  
Examples can be seen in Table \ref{analysis:case}. Given two sentences of text $s_1$ and $s_2$ , the models need to compute how similar $s_1$ and $s_2$  are, returning a similarity score between 0 and 5.  
As can be seen, scores from the proposed {\it surrogate} model are more correlated with to the gold compared to the universal sentence encoder and the SBERT model.

\begin{table}[t]
    \centering
    \scalebox{0.6}{
    \begin{tabular}{lp{9cm}r}\toprule
    {\bf Example 1} & & {\bf Score}  \\\midrule
     {\bf Sent 1:} & the problem likely will mean corrective changes & \textcolor{red}{4.4}  \\
    &  before the shuttle fleet starts flying again .  & \textcolor{blue}{0.74} \\
    {\bf Sent 2:} & he said the problem needs to be corrected  &  \textcolor{orange}{0.66} \\
    & before the space shuttle fleet is cleared to fly again .  &  \textcolor{ForestGreen}{0.43}   \\ \midrule
        {\bf Example 2} & & {\bf Score}  \\\midrule
     {\bf Sent 1:} &  every morning , they fly 240 miles to the farm . & \textcolor{red}{0.8}  \\
    &  & \textcolor{blue}{-0.74} \\
    {\bf Sent 2:} &  every morning , you fly 240 miles to every morning . &  \textcolor{orange}{-0.59} \\
    &   &  \textcolor{ForestGreen}{-0.13}   \\ \midrule
    {\bf Example 3} & & {\bf Score}  \\\midrule
     {\bf Sent 1:} & rt jones analyst juli niemann said grant was "the one& \textcolor{red}{1.4}  \\
    & we were all pulling for he has a very good reputation," & \textcolor{blue}{-0.71} \\
    {\bf Sent 2:} & rt jones analyst juli niemann said of grant . &  \textcolor{orange}{-0.39} \\
    &  &  \textcolor{ForestGreen}{0.19}   \\\midrule
      \bottomrule
    \end{tabular}
    }
    \caption{We use \textcolor{red}{gold}, \textcolor{blue}{surrogate},  \textcolor{orange}{sbert} and  \textcolor{ForestGreen}{universal}
    to denote scores obtained from the gold label,the proposed {\it Surrogate} model, the SBERT model\citep{reimers2019sentence} and the Universal Sentence Encoder model\citep{cer2018universal}, respectively. Scores from the proposed surrogate model are more correlated with to the gold compared to the universal sentence encoder and the SBERT model. 
   }  
    \label{analysis:case}
\end{table}

\section{Conclusion}
In this work, we propose a new framework for measuring sentence similarity   based on the fact that the probabilities of generating two similar sentences based on the same context should  be similar. We propose a pipelined system 
by first harvesting massive amounts of sentence pairs along with their similarity scores, and then training a surrogate model using the automatically labeled sentence pairs  for the purpose of faster inference. Extensive experiments demonstrate the effectiveness of the proposed framework against existing sentence embedding based methods.

\section*{Acknowledgement}
This work is supported by the Science and Technology Innovation 2030 - “New Generation Artificial Intelligence” Major Project (No. 2021ZD0110201) and the Key R \& D Projects of the Ministry of Science and Technology (2020YFC0832500).
We would like to thank editors for help and anonymous reviewers for their comments and suggestions. 

\bibliography{custom}
\bibliographystyle{acl_natbib}

\end{document}